\documentclass{article}

\usepackage[preprint,nonatbib]{nips_2018}

\usepackage[nonatbib]{nips_2018}
\usepackage{enumitem} 
\usepackage[utf8]{inputenc} 

\usepackage[T1]{fontenc}    
\usepackage{hyperref}       
\usepackage{url}            
\usepackage{booktabs}       
\usepackage{amsfonts}       
\usepackage{nicefrac}       
\usepackage{microtype}      
\usepackage{amsmath}
\usepackage{amsthm, amssymb, graphicx, tikz, hyperref, ifthen, xspace, bm, graphics}
\usepackage{fancyhdr}
\newcommand{\comment}[1]{}
\usepackage[square,sort,comma,numbers]{natbib}

\usepackage{fancyhdr}

\usepackage[nameinlink,noabbrev]{cleveref}
\newcommand*{\fullref}[1]{\hyperref[{#1}]{\Cref*{#1} -- \nameref*{#1}}}

\usepackage{enumerate}

\theoremstyle{plain}

\theoremstyle{definition}

\theoremstyle{remark}

\newcommand{\one}{1}
\newcommand{\indicator}[1]{\one_{\left\lbrace #1\right\rbrace}}
\DeclareRobustCommand\1{\futurelet\oneNext\oneCheck}%
\def\oneCheck{%
	\ifx\bgroup\oneNext \expandafter\indicator%
	\else%
	\expandafter\one%
	\fi%
}

\title{Understanding Gender and Racial Disparities in Image Recognition Models}

\author{
  Rohan Mahadev \\
  Department of Computer Science \\
  New York University \\
  New York, NY \\
  \texttt{rm5310@cs.nyu.edu} \\
  \And
  Anindya Chakravarti  \\
  Department of Computer Science \\
  New York University \\
  New York, NY \\
  \texttt{ac8184@nyu.edu} \\
}

\begin{document}

\maketitle

\section{Introduction and Background}

Modern computer vision has been one of the most widely used and significant applications of Deep Learning, which is predicated on the availability of two essential resources: 1: Clearly annotated large sets of data, and 2: Compute power capable of processing these large datasets in a relatively fast manner. With the advent of GPUs and subsequent advancements in being able to train deep neural networks on these GPUs, the second resource was in place. Thanks to the work by \citep{ILSVRC15}, the ImageNet dataset proved to be the final piece of the puzzle which led to the first successful use of Deep Learning to perform image classification \cite{krizhevsky2012imagenet}. Since then, systems performing tasks such as image classification and face recognition have been used widely to create employee attendance tracking systems to identifying suspects. Misidentification of people due to these systems can hence have adverse effects, such as being wrongly accused of a crime.   

\begin{figure}[h]
\begin{center}
\fbox{
\includegraphics[width=12cm,scale=1]{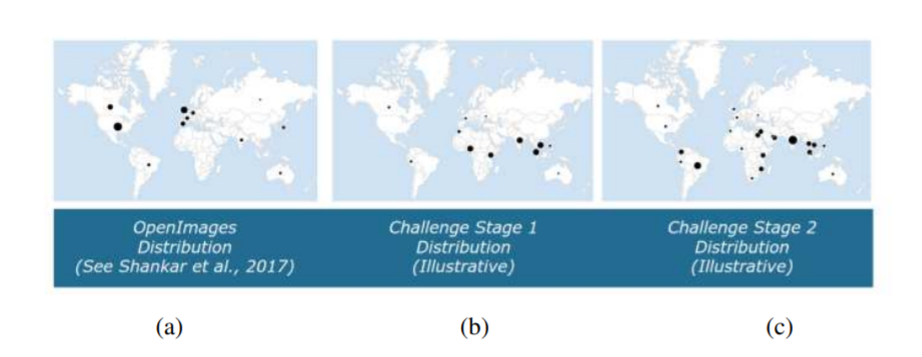}}
\end{center}
\caption{Geographical distribution of the Open Images dataset (a), and the evaluation sets for the Inclusive Images Challenge, (b) and (c)}
\label{fig:geog1}
\end{figure}

Large scale datasets such as ImageNet and Open Images \cite{OpenImages} are costly to create. In practice, using models pretrained on these datasets often perform better \cite{raghu2019transfusion} than training a model from scratch on a custom, smaller dataset. However, these datasets do not represent the real world scenario as can be seen from Fig. \ref{fig:geog1}.

An Automated Decision System (ADS) includes any technology that assists or replaces the judgement of human decision makers. In this report, we study and evaluate the 4th ranked solution of the Inclusive Images challenge on Kaggle. The main idea of the challenge is to develop models that do well at image classification tasks even when the data on which they are evaluated is drawn from a very different set of geographical locations than the data on which they are trained.

The Kaggle competition aims to reduce this disparity by using models which do well in the challenging area of distributional skew. Developing models and methods that are robust to distributional skew is one way to help develop models that may be more inclusive and fairer in real-world settings.

The data on which this \emph{fairer} model is evaluated can be found at \cite{Kaggle} and the model can be found at \cite{azatd}.

\section{Related Work}

The Gender Shades study \cite{GenderShades}, shows the disparity in the classification of three commercial gender classification algorithms tested on four subgroups of darker females, darker males, lighter females and lighter males. The datasets which are used to train these models are overwhelmingly composed of lighter-skinned subjects.
The study finds that the classifiers perform best for lighter individuals and males with up to 34\% disparity in misclassification between lighter and darker persons. The findings from this study provides the evidence for a need of increased demographic transparency in automated decision systems.

The study by Zou et al \cite{zou2018ai} gives an overview of a few AI applications that systematically discriminate against specific groups of population. Chen et al \cite{chen2018my} argue that the fairness of predictions should be evaluated in context of the data, and that unfairness induced by inadequate samples sizes should be addressed through data collection, rather than by constraining the model.

The Pew Research center conducted a study \cite{Pew} which shows the challenges of using machine learning to identify gender in images. Again, they found that every model was at least somewhat more accurate at identifying one gender than it was at the other – even though every model was trained on equal numbers of images of women and men. Crawford and Paglen \cite{Exc} also show the inherent biases in the machine learning training sets.

\section{Data Profiling}

The ADS was trained on the Google Open Images V6 dataset. Open Images is a dataset of ~9M images annotated with image-level labels. These images were collected by user submissions and manual labeling of images conducted by Google \cite{OpenImages}.

As per the holistic view of looking at data science, it is necessary to know where the data used in the ADS comes from. The geo-diversity analysis done by \citep{shankar2017no} shows that over 32\% of the data in the Open Images dataset originates from the United States and over 60\% of data originates from the six biggest countries in North America and Europe. On the flipside, China and India contribute to only 3\% of the dataset combined. The ImageNet dataset paints a similar picture.

\begin{figure}[h]
\begin{center}
\fbox{
\includegraphics[width=12cm,scale=1]{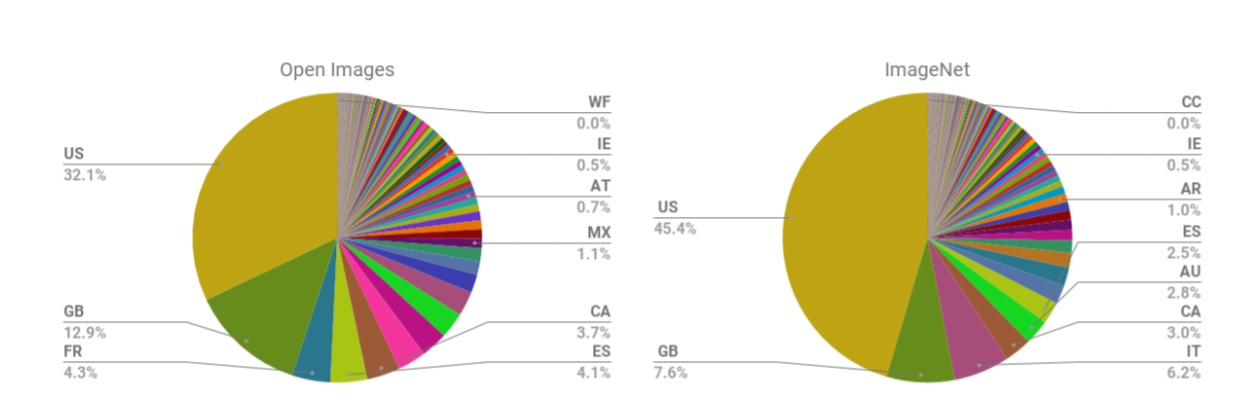}}
\end{center}
\caption{Countrywise geographical distribution of the Open Images dataset (a), and the ImageNet dataset (b)}
\label{fig:geog2}
\end{figure}

In our view, it is therefore essential to see how \emph{fair} these models are, as these models may be used in regions of the world which have close to no representation in the dataset.

\subsection{Input, Output and Interpretation}

\begin{figure}
\begin{center}
\fbox{
\includegraphics[width=12cm,scale=1]{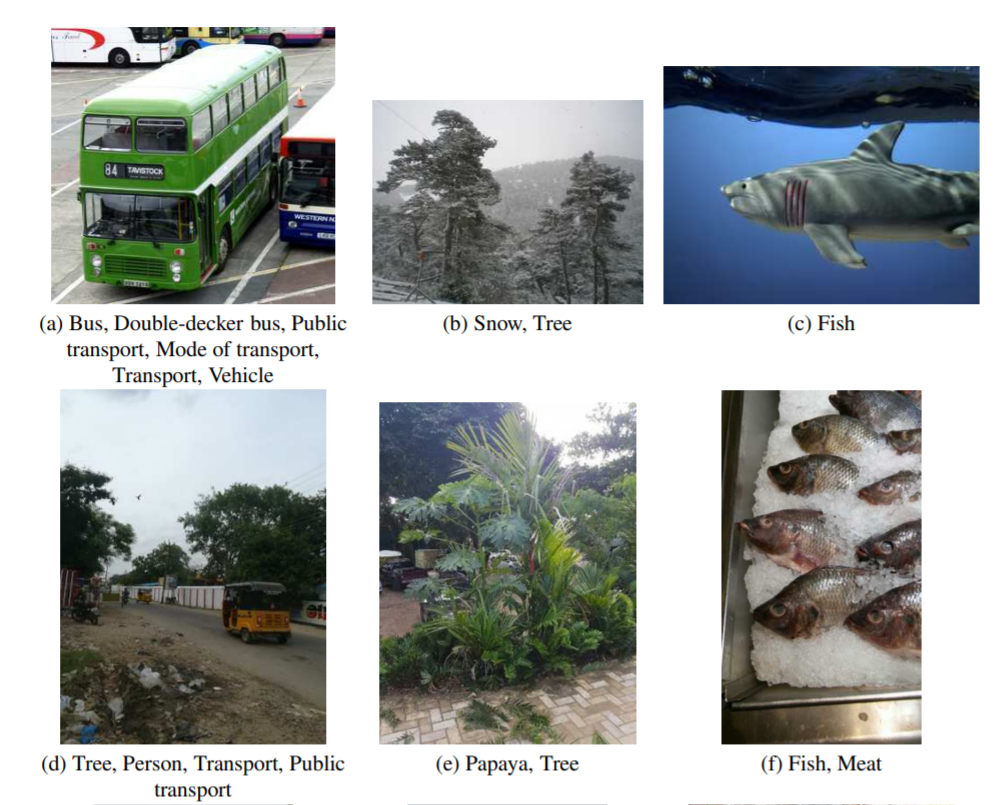}}
\end{center}
\caption{Images from the evaluation set of the Inclusive Images Challenge. These aren't useful to understand the fairness characteristics of the ADS.}
\label{fig:inclusive_eval}
\end{figure}

Studying the of the evaluation set of the Inclusive Images challenge, we found it to contain more noise than signal to be able to understand the fairness characteristics of the ADS, which can be seen from Figure \ref{fig:inclusive_eval}.

Hence, to understand the characteristics of this ADS, we chose to evaluate it over a sub-task of gender classification. The ADS takes in an image as an input and predicts categories to which the image may belong to, along with a confidence value. These categories could be one of 18 thousand different labels, so we only compare the confidence levels of the predictions of the labels "Man", "Woman", "Girl" and "Boy". For brevity, we combine the labels "Man" and "Boy" into "Male", and "Woman" and "Girl" to "Female".

So instead of using the Kaggle evaluation set, we use the MR2 dataset, \cite{MR2} which contains 74 images of men and women of European, African and East Asian descent to predict the gender of the people from the images. Using the race and gender of the subject in the image as protected attributes, we can understand the fairness metrics for the ADS. It is to be noted that the gender and race in the MR2 dataset are self-identified and are not crowd-sourced.

\begin{table}[h]
    \centering
    \begin{tabular}{|c|c|c|c|}
\hline
Sex & Race & N & Age \\
\hline
Female & African & 18 & 27.51(5.25) \\
& Asian & 12 & 25.17(4.73) \\
& European & 11 & 25.00(3.97) \\
\hline
Male & African & 14 & 27.20(5.27) \\
& Asian & 8 & 27.25(5.10) \\
& European & 11 & 26.69(3.78) \\
\hline
\end{tabular}
    \caption{Distribution of the MR2 dataset per sex and gender of subjects in the images}
    \label{tab:MR2}
\end{table}

\begin{figure}
\begin{center}
\fbox{
\includegraphics[width=8cm,scale=1]{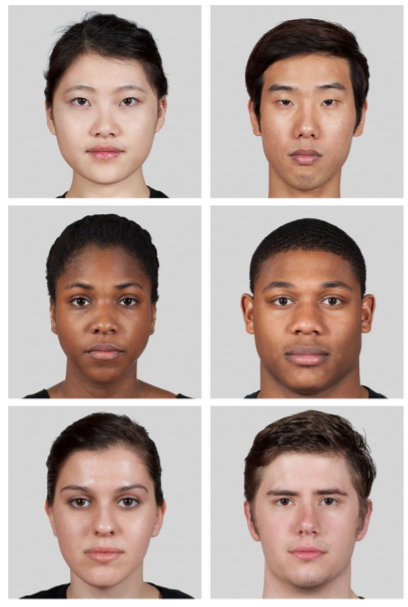}}
\end{center}
\caption{Sample images from each of the gender/race categories available in the MR2 database.}
\label{fig:mr2}
\end{figure}

\section{Implementation and Validation}
\subsection{ADS}
\par{Since the data we are dealing with are images, the pre-processing required is resizing of the images to make them compatible with the model and normalizing the image based on the std and mean of the training dataset.}

\par{
The ADS uses a Squeeze and Excitation Resnet (se\_resnet101) \cite{hu2018squeeze} which is a type of a Convolutional Neural Network. In order to mitigate the problem of distributional skew, the model uses a generalization of softmax with cross-entropy loss as opposed to the usual binary cross-entropy loss used in multi-label classification tasks. Further, the ADS uses the entirety of the Open Images dataset to train this model from scratch and uses random horizontal flips and crops to augment the dataset and provide regularization.
}
\par{
This ADS originally was tested on a test set provided on Kaggle which contained images from several geographical locations. Each image has multiple ground truth labels. We will use Mean F2 score to measure the algorithm quality. The metric is also known as the example based F-score with a beta of 2.
}
\par{
We however use the entirety of the MR2 dataset to predict the gender of the subjects in the image. We call a prediction to be a misclassification if the prediction gender does not match the self-identified gender in the dataset.
}

\subsection{Baseline Model}
\par{ 
To compare the difference made by the ADS model and to put the metrics into perspective, we tuned a baseline ResNet-18 model \cite{he2016deep} pretrained on ImageNet on CelebA \cite{liu2018large} dataset. The CelebA dataset contains more than 200K images of movie stars with 40 tagged attributes along with gender of the celebrity. This model achieves around 97\% accuracy on the test set of this dataset. 
}
\section{Outcomes}
\subsection{Accuracy and Performance}
We look at Accuracy metrics and performance of the baseline model and the ADS by considering the following protected attributes : race and sex. 
\subsubsection{Protected Attribute : Sex}
In this section, we discuss the inclusiveness of the models when predicting the Gender of a given image. We look at the entire dataset as a single group and look at the disparity in each of the subgroups (based on descent).

\begin{figure}[htp]
\centering
\includegraphics[width=.4\textwidth]{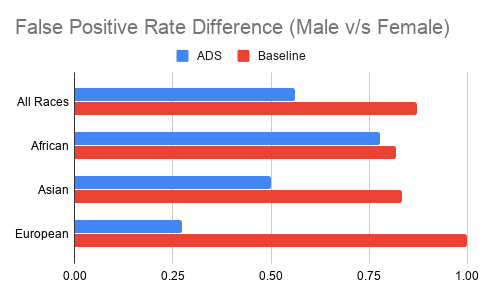}\quad
\includegraphics[width=.4\textwidth]{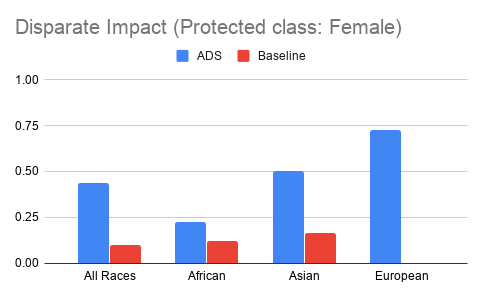}\quad

\medskip
\includegraphics[width=.4\textwidth]{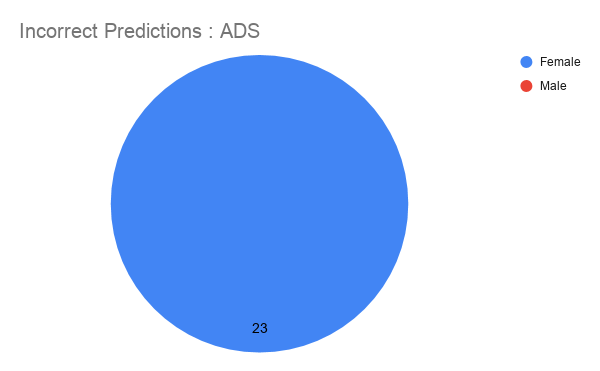}\quad
\includegraphics[width=.4\textwidth]{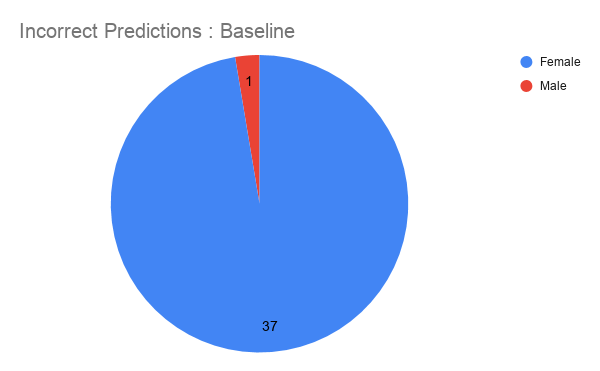}\quad

\caption{FPR Difference (top-left) and Disparate Impact(top-right) for ADS and Baseline model. Incorrect predictions for ADS and Baseline Model}
\label{fig:Nutrition_labels}
\end{figure}

\par{The two metrics that we used to compare the models were False Positive Rate (FPR) difference and Disparate Impact where true positive was considered when the gender was predicted correctly. The protected gender in this case was "Female". }

\par{
We can see from Fig. \ref{fig:Nutrition_labels} that the baseline model performs poorly across all the groups that we considered when compared to the ADS. Especially for "European" females where the baseline model predicted all of the images as males and therefore the higher FPR Difference and absence from Disparate Impact graph on the right. We can also see the ADS makes incorrect predictions for $23$ images tagged female while it makes zero incorrect predictions for male images. On the other hand, the baseline model performs poorly for both the sexes as well for the group on the whole.
}
\par{
Now, when looking at the performance of the ADS, we can see that it performs considerably well for "European" images while the performance on images tagged as "African"  is worse than the overall group. This is conforming with the fact that the Open Images dataset on which the model was trained and validated contains images taken primarily from North American and European countries. 
}

\subsubsection{Protected Attribute : Race}
Based on the analysis in the previous section, we know the ADS performs well for Male subset of the data while it has poor performance for the female subset of the data. Now, we look how the model performs for different races and look at the same fairness metrics for each.

For this analysis, we use our knowledge of the Open Images dataset distribution to define "European" class as the privileged class and look at the other two races (Asian and African) in the dataset as protected groups one at a time. The figure below shows the performance measures.

\begin{figure}[htp]
\centering
\includegraphics[width=0.65\textwidth]{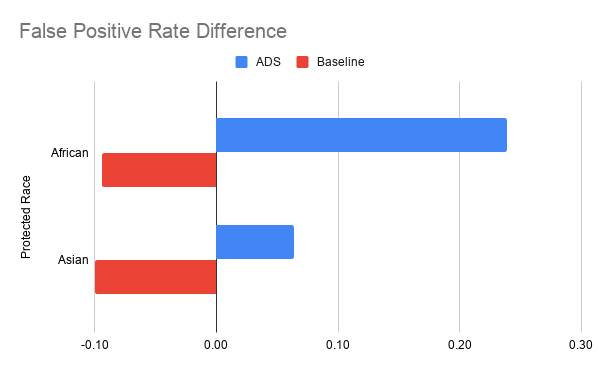}\quad

\caption{FPR Difference with protected attribute, race, for ADS and Baseline model.}
\label{fig:FPR_Race}
\end{figure}

\par{
We define True Positive as the number of correct predictions for a given race. We can clearly see form Fig. \ref{fig:FPR_Race} that Baseline model performs poorly for the privileged class as compared to the protected classes. But the ADS still performs worse for protected classes when compared to the privileged class. In addition to this, the ADS also performs poorly for images tagged as "African" while it works comparatively well on "Asian" images where the FPR difference is $0.238$ for the former and $0.064$ for the latter.
}

\begin{figure}[htp]
\centering
\includegraphics[width=0.65\textwidth]{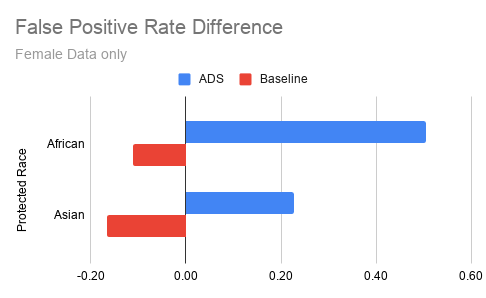}\quad

\caption{FPR Difference with protected attribute, race, for ADS and Baseline model. Female Data only.}
\label{fig:FPR_Race_female}
\end{figure}

\par{
Deriving from the conclusions of the previous two discussions, we now consider a subset of the data containing only images tagged as Female and measure fairness for different protected races. The expectation is that the fairness metrics would follow the explanation from the section above and the ADS would perform worse for images tagged to race "African". This is because the other subset of the data (containing the Male images) has $100\%$ accuracy for all the races. The figure below supports this hypothesis.

We do not compare the Disparate Impact in this section because FPR for privileged class is 1 and hence the DI value is $\infty$.
}
\subsection{Interpreting the ADS}

Understanding the reasons behind why a machine learning model is important in assessing trust, which is of the essence if a model is to be deployed for public use. To do so, we use the LIME technique which explains the predictions of our classifier by learning an interpretable model locally around the prediction \cite{lime}. LIME highlights pixels in an image to give an intuition as to why the model thinks that a certain class may be present in the image.

\begin{figure}[htp]
\centering
\includegraphics[width=.4\textwidth]{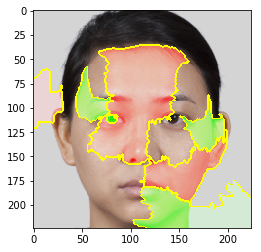}\quad
\includegraphics[width=.4\textwidth]{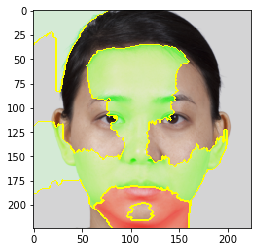}\quad

\medskip
\includegraphics[width=.4\textwidth]{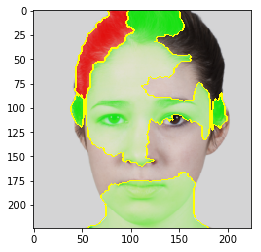}\quad
\includegraphics[width=.4\textwidth]{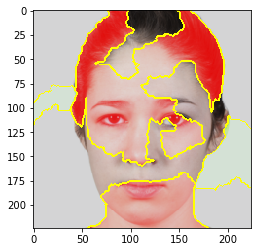}\quad
\medskip
\includegraphics[width=.4\textwidth]{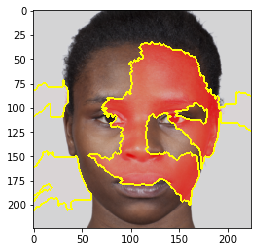}\quad
\includegraphics[width=.4\textwidth]{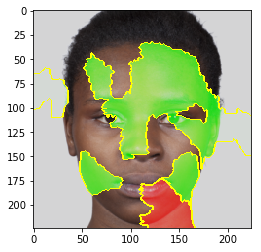}\quad

\caption{LIME explanations for the ADS for females belonging to different races from the MR2 dataset. Explanation for prediction of the "Female" class (left) and explanation for prediction for the "Male" class (right). Correct prediction : row 2.}
\label{fig:LimeExplanations}
\end{figure}

We try to interpret the explanation generated by LIME for the predictions done by the ADS. The only correct prediction in Fig. \ref{fig:LimeExplanations} is in the second row where a European female is correctly classified. By look at the other explanations of similar correct predictions, we believe that the classifier is learning very general female attributes around the eyes and chin. At the same time, it seems that the classifier is looking at the cheekbones and the background in the images (see the green patch in the background of row 1 and 3 Male prediction explanation) to classify as image as a Male. This green patch was present as contributing to Male classification in many of the images (correct and incorrect predictions). It was also observed that the classifier was trying to use the hair texture to classify an image but we couldn't be sure of what and how it was using this.

\section{Summary}
There is certainly a long way to go for general image classifiers to do Image Inclusiveness across all genders (more than binary) across all races. For this, we certainly need more balanced and robust datasets. The stakeholders who would benefit the most with the current set of fairness metrics would be commercial face recognition services such as the ones from IBM and Face++. Having said that, we believe that the ADS improved certain aspects of the classification as is evident from the study of the metrics and it's comparison with a naive base model (with respect to the task) trained to classify mundane objects. This is surely a good step towards more generic, accurate and fair models. Challenges like this, from large corporations like Google certainly help in this regard and we can hope that the models will become more robust in the future.

\section{Clarifications}
We would like to state we present these findings not as a criticism of the ADS but as a case-study in the difficulty of solving the problem of distributional skew. The Inclusive Images Challenge clearly states that the winning solutions are not necessarily \emph{fair} in all aspects. There are a wide variety of definitions of fairness and we chose one such measure based on gender identities. 

\newpage
\bibliographystyle{plain}
\bibliography{main}

\begin{thebibliography}{10}

\bibitem{krizhevsky2012imagenet}
Krizhevsky Alex, Sutskever Ilya, and Geoffrey~E Hinton.
\newblock Imagenet classification with deep convolutional neural networks.
\newblock In {\em Advances in neural information processing systems}, pages
  1097--1105, 2012.

\bibitem{chen2018my}
Irene Chen, Fredrik~D Johansson, and David Sontag.
\newblock Why is my classifier discriminatory?
\newblock In {\em Advances in Neural Information Processing Systems}, pages
  3539--3550, 2018.

\bibitem{Exc}
Kate Crawford and Trevor Paglen.
\newblock The politics of images in machine learning training sets.
\newblock \url{https://www.excavating.ai/}.
\newblock Accessed: 2020-03-30.

\bibitem{azatd}
Azat Davletshin.
\newblock 4th place solution - inclusive-images-challenge.
\newblock \url{https://github.com/azat-d/inclusive-images-challenge}, 2019.

\bibitem{GenderShades}
Buolamwini~J. Gebru~T.
\newblock Gender shades: Intersectional accuracy disparities in commercial
  gender classification.
\newblock {\em MIT Media Lab}, 2018.

\bibitem{hu2018squeeze}
Jie Hu, Li~Shen, and Gang Sun.
\newblock Squeeze-and-excitation networks.
\newblock In {\em Proceedings of the IEEE conference on computer vision and
  pattern recognition}, pages 7132--7141, 2018.

\bibitem{he2016deep}
He~Kaiming, Zhang Xiangyu, Ren Shaoqing, and Jian Sun.
\newblock Deep residual learning for image recognition.
\newblock In {\em Proceedings of the IEEE conference on computer vision and
  pattern recognition}, pages 770--778, 2016.

\bibitem{OpenImages}
Alina Kuznetsova, Hassan Rom, Neil Alldrin, Jasper Uijlings, Ivan Krasin, Jordi
  Pont-Tuset, Shahab Kamali, Stefan Popov, Matteo Malloci, Alexander
  Kolesnikov, Tom Duerig, and Vittorio Ferrari.
\newblock The open images dataset v4: Unified image classification, object
  detection, and visual relationship detection at scale.
\newblock {\em IJCV}, 2020.

\bibitem{liu2018large}
Ziwei Liu, Ping Luo, Xiaogang Wang, and Xiaoou Tang.
\newblock Large-scale celebfaces attributes (celeba) dataset.
\newblock {\em Retrieved August}, 15:2018, 2018.

\bibitem{raghu2019transfusion}
Maithra Raghu, Chiyuan Zhang, Jon Kleinberg, and Samy Bengio.
\newblock Transfusion: Understanding transfer learning for medical imaging.
\newblock In {\em Advances in Neural Information Processing Systems}, pages
  3342--3352, 2019.

\bibitem{Kaggle}
Google Research.
\newblock Inclusive images challenge.
\newblock \url{https://www.kaggle.com/c/inclusive-images-challenge/}.

\bibitem{lime}
Marco~Tulio Ribeiro, Sameer Singh, and Carlos Guestrin.
\newblock "why should {I} trust you?": Explaining the predictions of any
  classifier.
\newblock In {\em Proceedings of the 22nd {ACM} {SIGKDD} International
  Conference on Knowledge Discovery and Data Mining, San Francisco, CA, USA,
  August 13-17, 2016}, pages 1135--1144, 2016.

\bibitem{ILSVRC15}
Olga Russakovsky, Jia Deng, Hao Su, Jonathan Krause, Sanjeev Satheesh, Sean Ma,
  Zhiheng Huang, Andrej Karpathy, Aditya Khosla, Michael Bernstein,
  Alexander~C. Berg, and Li~Fei-Fei.
\newblock {ImageNet Large Scale Visual Recognition Challenge}.
\newblock {\em International Journal of Computer Vision (IJCV)},
  115(3):211--252, 2015.

\bibitem{shankar2017no}
Shreya Shankar, Yoni Halpern, Eric Breck, James Atwood, Jimbo Wilson, and
  D~Sculley.
\newblock No classification without representation: Assessing geodiversity
  issues in open data sets for the developing world.
\newblock {\em arXiv preprint arXiv:1711.08536}, 2017.

\bibitem{MR2}
Nina Strohminger, Kurt Gray, Vladimir Chituc, Joseph Heffner, Chelsea Schein,
  and Titus~Brooks Heagins.
\newblock The mr2: A multi-racial, mega-resolution database of facial stimuli.
\newblock {\em Behavior research methods}, 48(3):1197--1204, 2016.

\bibitem{Pew}
Stefan Wojcik and Emma Remy.
\newblock The challenges of using machine learning to identify gender in
  images.
\newblock
  \url{https://www.pewresearch.org/internet/2019/09/05/the-challenges-of-using-machine-learning-to-identify-gender-in-images/}.
\newblock Accessed: 2020-03-30.

\bibitem{zou2018ai}
James Zou and Londa Schiebinger.
\newblock Ai can be sexist and racist—it’s time to make it fair, 2018.

\end{thebibliography}

\end{document}